# Flaws of ImageNet, Computer Vision's Favorite Dataset

Since its release, ImageNet-1k dataset has become a gold standard for evaluating model performance. It has served as the foundation for numerous other datasets and training tasks in computer vision.
As models have improved in accuracy, issues related to label correctness have become increasingly apparent. In this blog post, we analyze the issues in the ImageNet-1k dataset, including incorrect labels, overlapping or ambiguous class definitions, training-evaluation domain shifts, and image duplicates. The solutions for some problems are straightforward. For others, we hope to start a broader conversation about refining this influential dataset to better serve future research.

| AUTHORS | AFFILIATIONS |
|---|---|
| **Nikita Kisel** | Visual Recognition Group, Czech Technical University in Prague |
| **Illia Volkov** | Visual Recognition Group, Czech Technical University in Prague |
| **Kateřina Hanzelková** | Faculty of Science, Charles University |
| **Klara Janouskova*** | Visual Recognition Group, Czech Technical University in Prague |
| **Jiri Matas** | Visual Recognition Group, Czech Technical University in Prague |

**\*Corresponding author:** janoukl1@fel.cvut.cz

## Disclaimer:

By undertaking this work, we have no intention to diminish the significant contributions of the ImageNet, whose value remains undeniable. It was, at the time of its publication, far ahead of all existing datasets. Given ImageNet-1k's continued broad use, especially in model evaluation, fixing the issues may help the field move forward. With current tools, we belive it is possible to improve ImageNet-1k without huge manual effort.

# Introduction to ImageNet-1k

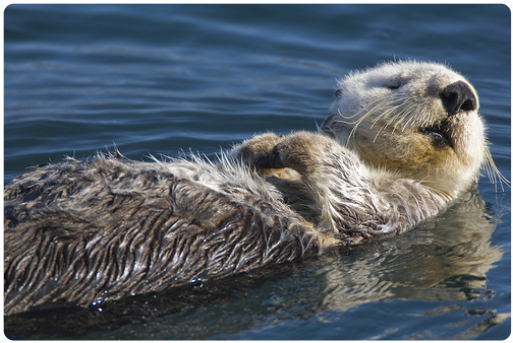
(a) *"otter"* ✓

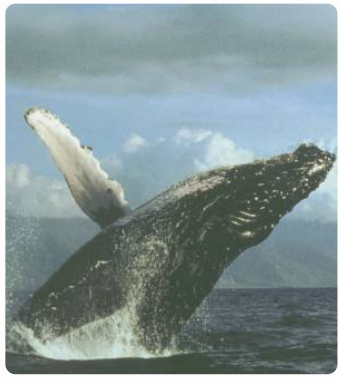
(b) *"tiger shark"* ✗
*"grey whale"* ✓

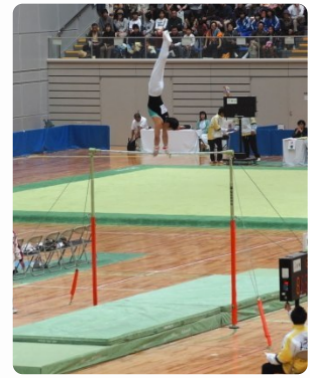
(c) *"parallel bars"* ✗
*"horizontal bar"* ✓

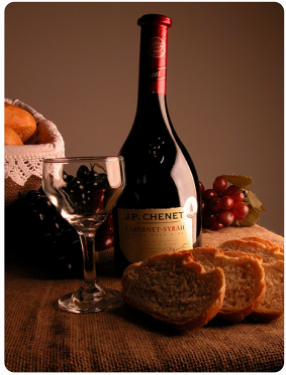
(d) *"wine bottle"* ✓
+ *"goblet"* ✓
+ *"red wine"* ✓

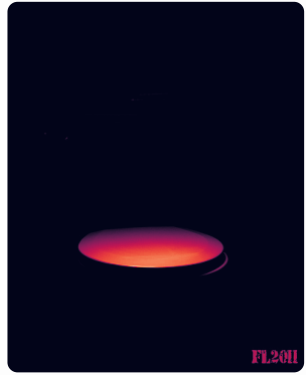
(e) *"toilet seat"* ✗

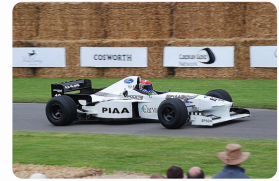
(f) *"race car"* ✓

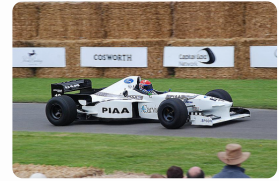
(g) *"sports car"* ✗
*"race car"* ✓

Figure 1. Examples of ImageNet issues. The top label is the ground truth; colors indicate **correct** ✓; and **incorrect** ✗ labels. When the ground truth is wrong, the correct label is below. Labels of objects from other ImageNet classes that are present in the image are marked "+". (a), (f) *Correctly labeled images*. (b), (c), (e), (g) *Incorrectly labeled images*. (d) *Images with multiple objects*. (e) *Ambiguous images*. (f), (g) *Pixel-wise duplicates* (the top image is from the training, the bottom from the validation set).

**Brief History**

The concept of **ImageNet** [1] was introduced in 2009. It was to become the first large-scale labeled image dataset, marking a transformative moment in computer vision. In its construction, the authors followed the structure of WordNet [2] — a lexical database that organizes words into a semantic hierarchy.
With over 3 million labeled images from more than 5,000 categories in 2009, it far surpassed previous commonly used datasets, which contained only tens of thousands of images (e.g., LabelMe with 37,000 images, PASCAL VOC with 30,000 images, CalTech101 with 9,000 images).
This unprecedented scale allowed ImageNet to capture a diverse range of real-world objects and scenes. As of November 2024, the ImageNet [1:1] paper has since been cited over 76,000 times according to Google Scholar and more than 34,000 times in IEEE Xplore, underscoring its profound impact on computer vision. Its influence is so significant that

even in other fields of machine learning, the expression
"ImageNet moment" is used to describe groundbreaking developments.

Introduced in 2012, **ImageNet-1k** [3] was created specifically for the ImageNet Large Scale Visual Recognition Challenge (ILSVRC). It contains 1,000 diverse classes of animals, plants, foods, tools, and environments (e.g., lakesides, valleys). It includes a training set of over 1 million images from the original ImageNet, along with validation and test sets of 50,000 and 100,000 images, respectively. The evaluation subsets were formed by a process closely following the original dataset creation. It is important to note that, since the test set labels are not publicly available, the validation set is typically used for model evaluation.

**Problems**

We were aware that ImageNet-1k had issues, as is common in nearly every human-annotated real-world dataset. However, while analyzing model errors in another project, we were surprized by the size of the effects of the problems related to the ground truth. We decided to investigate them in greater detail. Our initial goal was simple: fix the labels to reduce noise — a task we assumed would be straightforward. However, upon further examination, we discovered that the issues were complex and far more deeply rooted than expected.

Some of the issues are easier to solve, such as incorrect image labels, overlapping classes, and duplicate images. These can be addressed by relabeling the images where possible, removing irrelevant or dependent classes, and cleaning up duplicates. However, there are other issues, such as a distribution shift between the training and evaluation subsets, where we see no effective solution; it could be even seen as a feature rather than a bug since in real deployment, test and training data are never i.i.d. (independent identically distributed).
The inclusion of multilables images depicting multiple objects from different ImageNet classes poses another problem with no obvious, backward compatible solution. The removal of such images is a possibility when ImageNet-1k is used for training. However, their prevalence, which we estimate to be in range from 15% to 21% in the validation set, probably rules out adopting this solution for standard ImageNet-1k evaluation. Modifications of the evaluation protocol might be needed, e.g. considering any label corresponding to a depicted object as a correct output.

# Known ImageNet-1k Issues

Prior studies identified and analyzed issues related to the ImageNet dataset, but they each deal only with a specific concern and usually focus on the validation set. The topic that received most attention are annotation errors, which distort the evaluation of model accuracy [4][5][6][7][8].

Overlapping class definitions were reported in Vasudevan et al. [9]. Duplicate images, which cause overestimation of model performance for certain classes, were mentioned in several papers [9:1][10][11].

For more information on each work individually, see Appendix A.1.

## Bringing the Errors Together

After downloading the available error analysis from prior works, we discovered that approximately 57.2% of the images in the validation set were reviewed by multiple studies. Further examination revealed that only 33% of the entire validation set had identical labels across all the studies that reviewed the images. This finding reminded us that error-correction processes are not error-free. We analyzed the images with consistent labels, and for nearly 94% of them the original labels were correct. The remaining 6% consisted of images where the original label was incorrect but had full agreement on the corrected label, as well as images that were either multi-labeled or ambiguous. For more details about previous work evaluation, see Appendix A.2.

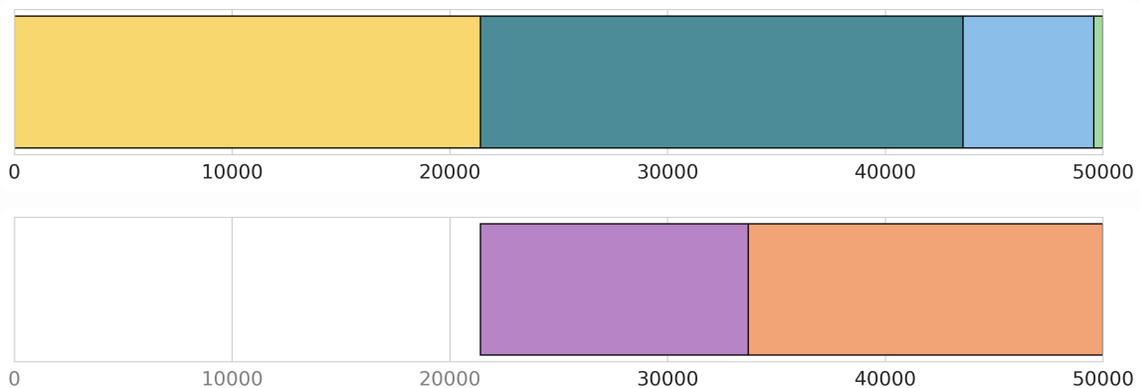

Figure 2. Top: Number of images checked by one ■, two ■, three ■ and four ■ papers.
Bottom: Images checked by more than one paper where the annotators agreed ■ and disagreed ■.

This analysis highlights significant inconsistencies in the labeling of reannotated images across different studies. These discrepancies arise because each study followed its own methodology, leading to varying interpretations of the class definitions and different approaches to resolving issues encountered during annotation. How can one accurately annotate images for an ambiguous class without first conducting a thorough analysis of the class definitions? We will explore this question in greater detail in the case study section.

# Dataset Construction Issues

Let us first examine the two-step process used to construct ImageNet:

1. **Image collection.** Images were scraped from the web and organized according to the WordNet hierarchy, which will be revisited later. Both automated and manual methods were involved: automated tools initially assigned labels based on information available (textual description, category) at the image source, often [flickr](#) or general web search.
2. **Annotation process**. The preliminary labels were then reviewed by MTurk workers, who were only asked to confirm whether the object with the given label was present in the image. Notably, no alternative labels were suggested, such as other similar classes. The Mturkers were shown the target synset definition and a link to Wikipedia.

**ImageNet-1k** was constructed later. It consists of three sets: a *training set* with over 1 million images, a *validation set* with 50,000 images, and a *test set* with 100,000 images. The training set is drawn from the ImageNet. Images for the evaluation subsets were obtained through a process that tried to replicate the one used for the ImageNet. The new data were collected up to three years later than training data and then they were randomly split between the validation and test sets.

ImageNet links each image category to a specific noun **WordNet** synset. WordNet is a comprehensive lexical database of English. It organizes nouns, verbs, adjectives, and adverbs into cognitive synonym sets, or synsets, each representing a unique concept (see the official [website](#)).

Each **synset** consists of one or more terms referred to as synonyms, for example *"church, church building"*. However, this is not true in all cases. For example, consider the synset *"diaper, nappy, napkin"*. Even though the first terms are synonyms, the third one is not. Moreover, there are cases where the same term belongs to more than one synset, e.g. there are two synsets named *"crane"* — one defining a bird and the second a machine. In ImageNet-1k they are separate classes. Think about the consequences for zero-shot classification with vision-language models (VLMs) like CLIP [12].

We will demonstrate some issues related to dataset construction on a couple of examples.

## The Cat Problem

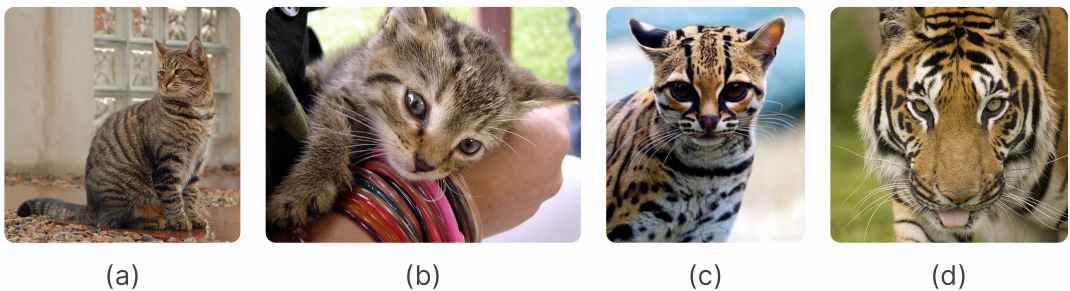

(a)    (b)    (c)    (d)

Figure 3. Images from the *"tiger cat"* class. (a) is a tiger cat. (b) is a cat, but the coat pattern is not clearly visible, it might be a *"tabby, tabby cat"*, another ImageNet-1k class. (c) is a wild cat from the *Leopardus* genus, not a domestic cat. (d) is a tiger.

Looking at the *"tiger cat"* images above, you might think "This seems to be a really diverse dataset...". Let us have a closer look. The *"tiger cat"* class is defined in WordNet as **"a cat having a striped coat"**, which aligns precisely with (a).

To understand the issue with (b), we must know that ImageNet-1k also includes a *"tabby, tabby cat"* class, defined as **"a cat with a grey or tawny coat mottled with black"**. In common usage, tabby cat refers broadly to any domestic cat with a striped, spotted, or swirled coat pattern, all of which must include an "M" marking on their forehead (which can clearly be seen on this image). Most dictionaries agree that all tiger cats are tabbies, but not all tabby cats have the tiger pattern. However, even if we look at the image (b) through the lens of WordNet definitions, it shows a grey cat, but its coat isn't clearly visible. Moreover, the term "mottled coat" in the tabby cat definition can be somewhat confusing, as some dictionaries consider stripes to be a type of mottling. So, how do we determine which type of cat this is?

We find modern large language models (LLMs) to be more accurate when handling such questions, so we asked them whether these two definitions overlap:

> *Yes, the "tabby, tabby cat" definition and the "tiger cat" definition overlap. While the first definition is broader in its description of coloration and pattern, the second one specifies the striped aspect, which is a common characteristic of the broader "mottled" description in the first.*

– ChatGPT-4o

> *Yes - a tabby cat definition ("a cat with a grey or tawny coat mottled with black") overlaps with the tiger cat definition ("a cat having a striped coat") since "mottled with black" typically manifests as stripes in domestic cats.*

– Claude 3.5 Sonnet

> *Yes, mottled patterns can sometimes include stripes, so the "tabby, tabby cat" definition can occasionally overlap with the "tiger cat" definition.*

– Microsoft Copilot

This raises the question: *If WordNet definitions are not precise enough, what is the ultimate source for correct image labels? ChatGPT, Wikipedia, GBIF?*

We are not wildlife experts, but we can say that either an oncilla, ocelot, or margay may be seen in (c). While this might seem like harmless noise, common in such large datasets, these animals do appear more than once in the training set. In everyday language, *"tiger cat"* is even more commonly used to refer to these wild cats than to a striped domestic ones; however, these usages coexist simultaneously.

We have already mentioned the WordNet definition of the *"tiger cat"* synset; WordNet also contains ***"tiger cat, Felis tigrina"*** synset, defined as **"a medium-sized wildcat of Central America and South America having a dark-striped coat"**.

All three of the possible species of cats we've mentioned as possible labels for (c) fall under this definition. Consistently annotating *"tiger cat"* images given such confusing background is difficult for experts, and probably impossible for MTurkers.

Obviously, (d) is a tiger, which has its own synset, ***"tiger, Panthera tigris"***, in ImageNet-1k. Such tigers make up a significant portion of the "tiger cat" class in both the training and validation sets.

Distinguishing between a tabby and a tiger isn't even a particularly challenging fine-grained recognition task. While using non-expert annotators can be cost-effective and quick, this example highlights the need to think carefully about the expertise and motivation of those labeling the data.

## The Laptop Problem

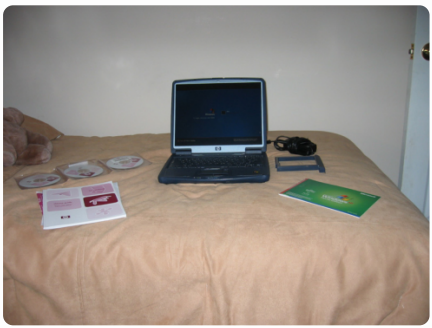
*"laptop, laptop computer"*

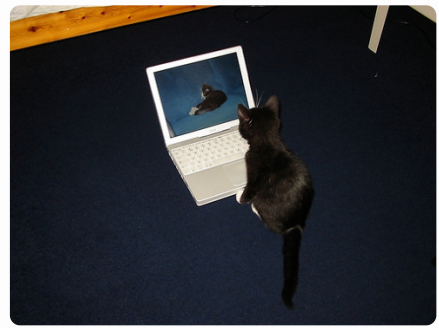
*"notebook, notebook computer"*

**Figure 4. The terms laptop and notebook are now used interchangeably. Black cat, unlike *"tabby cat"* and *"tiger cat"*, is not an ImageNet-1k class.**

Consider two synsets: *"laptop, laptop computer"* and *"notebook, notebook computer"*. Their respective WordNet definitions are "a portable computer small enough to use in your lap" and "a small compact portable computer". In this case, the definitions clearly overlap, with the first being a subset of the second. We again asked modern LLMs about the difference between laptops and notebooks:

> *Previously, "laptops" referred to portable computers with larger screens and more powerful hardware, while "notebooks" were slimmer, lighter, and less powerful. Today, the terms are used interchangeably, as advancements in technology have blurred the distinctions.*

– *ChatGPT-4o*

This raises a question whether there is any other solution besides merging these two classes into a single class with 2600 training images or changing the evaluation protocol so that laptop – notebook swaps are not penalized.

## Exploring VLM Results

We expected the issues described above to have a clear impact on results of Vision Language Models. To test this hypothesis, we selected a zero-shot open-source model, OpenCLIP (ViT-H-14-378), and examined its predictions on the training set for the classes discussed above. The confusion matrices below show the discrepancies between the original and predicted labels.

| Original Label ↓ | Predicted Label (OpenCLIP) → | | | |
| --- | --- | --- | --- | --- |
| | tabby, tabby cat | tiger cat | tiger, Panthera tigris | Other classes |
| tabby, tabby cat | 76.4% | 8.5% | 0% | 15.1% |
| tiger cat | 57.2% | 6.9% | 23.8% | 12.1% |
| tiger, Panthera tigris | 0.1% | 0% | 99.2% | 0.7% |

Table 1. OpenCLIP predictions for classes related to 'The Cat Problem'.

Note that the differences can be both due to OpenCLIP's errors and wrong ground truth labels. Nearly a quarter of the images in *"tiger cat"* class are predicted to be tigers, which we trust to be an estimate of the percentage of tigers in the training data of the class. Only 6.9% of images are predicted as *"tiger cat"*, highlighting the conceptual overlap with *"tabby, tabby cat"*.

| Original Label ↓ | Predicted Label (OpenCLIP) → | | |
| --- | --- | --- | --- |
| | laptop, laptop computer | notebook, notebook computer | Other classes |
| laptop, laptop computer | 35.8% | 44.6% | 19.6% |
| notebook, notebook computer | 17.2% | 65.8% | 17% |

Table 2. OpenCLIP predictions for classes related to 'The Laptop Problem'.

Approximately 80% of the images in both classes were predicted to be either a notebook or a laptop, with an error not far from random guessing. The remaining 20% were assigned to other labels. This interesting observation will be discussed in the section on multilabels.

## Key Takeaways

The examples demonstrate that the incorrect labels are not just random noise, but are also an outcome of the dataset's construction process. WordNet might not have been the most suitable foundation to build on, as its definitions are not precise enough. Also, some meanings shift over time, which is a problem in the era of VLMs. Perhaps WordNet and ImageNet should co-evolve.

Relying solely on MTurkers and using Wikipedia (a source that may be edited by non-experts, updated in real time, or lack precise definitions) not only led to the inclusion of noisy labels but also sometimes distorted the very concepts that the classes were intended to represent. For example, the *"sunglasses, dark glasses, shades"* and *"sunglass"* classes represent the same object — sunglasses. While this is accurate for the former class, the latter class is defined in WordNet as "a convex lens that focuses the rays of the sun; used to start a fire".
This definition was lost during the dataset's construction process, resulting in two classes representing the same concept.

## Distribution Shift Between Training and Validation Sets

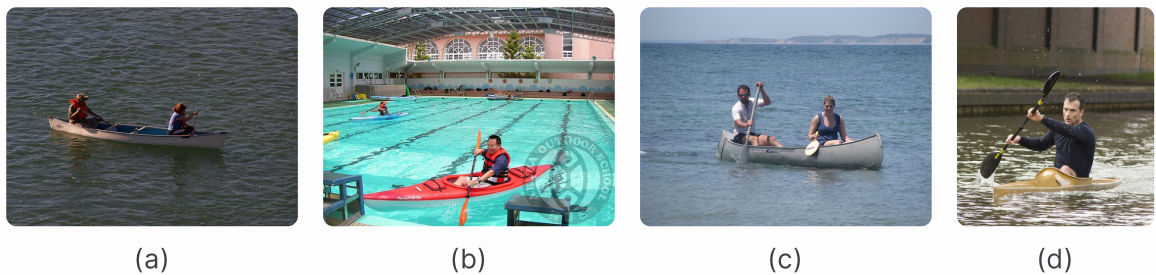

(a) (b) (c) (d)

Figure 5. Distribution shift between training and validation sets. (a) *"canoe"* in the training set (actual canoe). (b) *"canoe"* in the validation set (a kayak). (c) *"paddle, boat paddle"* with a presence of a canoe. (d) *"paddle, boat paddle"* with a presence of a kayak.

As mentioned earlier, for ImageNet-1k, additional images were collected using the same strategy as the original ImageNet. However, even with the same process, issues arose.

For example, in the training set, the *"canoe"* class mainly consists of images of canoes, but it also includes many images of kayaks and other types of boats. In contrast, **the *"canoe"* class in the validation set only contains images of kayaks, with no canoes at all**.

To clarify, the difference is not only in the boat shapes, with a kayak being more flat, but also in how they are paddled. A canoe is typically paddled in a kneeling position (though seated paddling is common) with a short single-bladed paddle, while a kayak is paddled from a seated position with a long double-bladed paddle. Interestingly, "paddle, boat paddle" is also a separate class in ImageNet-1k.

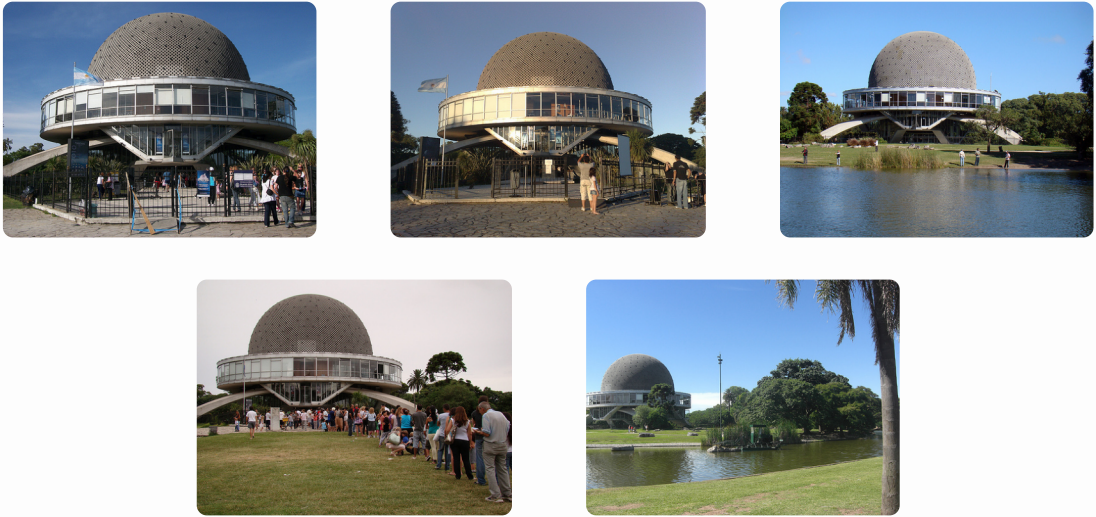

**Figure 6.** Images from the *"planetarium"* class in the ImageNet validation set. 68% of validation images in this class depict this particular building.

The *"planetarium"* class exibits another issue. In the validation set, 68% of the images (34 out of 50) feature the same planetarium in Buenos Aires. The building appears many times in the training set too, but there the range of planetaria is much broader. Perhaps it is a beautiful location, and the authors enjoyed featuring it, but it is clear not i.i.d. to have this one appear so frequently in the validation set.

Here, the solution is fairly straightforward - a more representative set of images can be collected.
Nevertheless, this breaks backward compatability, rendering old and new results incomparable.

## Images Deserving Multiple Labels

We will illustrate the problem with multilabel images with an extreme example. What do you think should be the correct label for the following image?

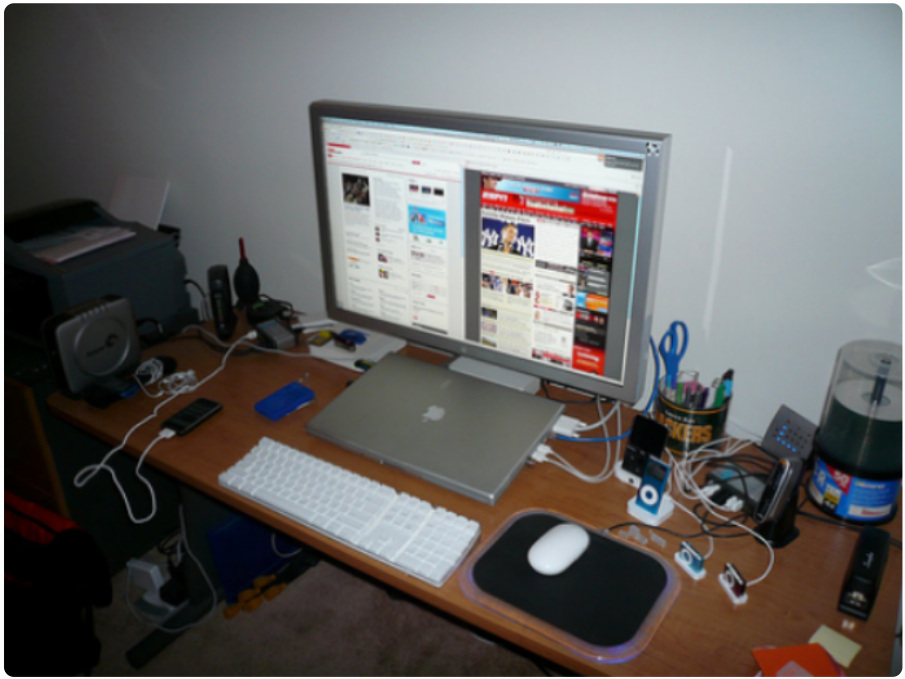

**Figure 7. An image from the** *"mouse, computer mouse"* **class. Objects from at least 11 other ImageNet-1k classes are visible.**

All of the following objects in the image have their own class in ImageNet-1k:

- Computer keyboard
- Space bar
- Monitor
- Screen
- Notebook
- Laptop
- Desktop computer
- Desk
- iPod
- Website
- Printer

One could argue that in the presence of multiple objects from distinct classes, the dominant should be labeled. This image shows it is often not clear what the dominant object is.

As with the *"canoe"* and *"paddle"* classes in the section about domain shift, some objects naturally appear together in photos. In everyday usage, desktop computers are accompanied by a computer keyboard and a monitor (all of which are ImageNet-1k classes). The difference between a monitor and a screen, yet another set of questionable ImageNet-1k classes, is an interesting question in its own right. Additionally, desktop computers are generally placed on desks (also a class), so these two objects often appear together in images. Many such cases of multilabel issues stemming from frequently co-occurring classes exist.

After careful examination, the issue runs deeper and the authors' claim that there is no overlap and no parent-child relationship between classes appears to be incorrect. Consider the example of a spacebar and a computer keyboard. The space bar may not always be part of a computer keyboard, but most keyboards do have a space bar.

Let us look at another example to further explore the topic.

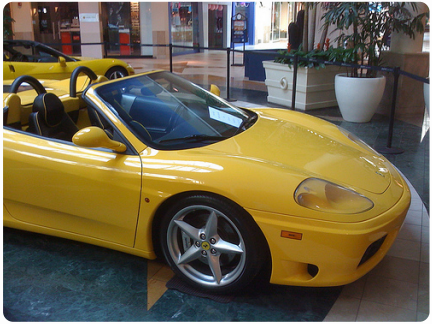 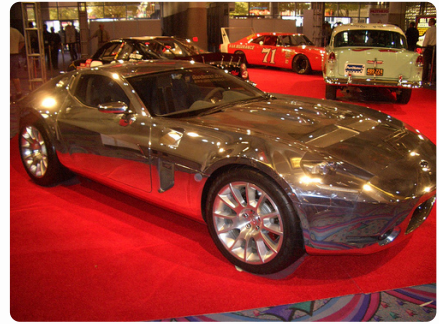

*"car wheel"*            *"sports car, sport car"*

Figure 8. *"car wheel"* and *"sports car, sport car"*. There are not many cars without a wheel.

A wheel can exist without a car, but a car — except for some rare cases, say in a scrapyard — cannot exist without a wheel. When an image contains both (and many do), it becomes unclear which label should take priority. Even if MTurkers were familiar with all 1000 ImageNet-1k classes, assigning a single accurate label would still be challenging.

As mentioned in the section about dataset construction, MTurkers were asked to determine whether an image contains an object that matches a given definition. Such a process of annotation may not be inherently problematic. However, when paired with the problematic class selection, our next topic, it is.

## ILSVRC Class Selection

The ImageNet-1k classes were chosen as a subset of the larger ImageNet dataset. One reason the dataset is so widely used is that it is perceived to reflect the diversity of the real world. The class distribution is distorted; does having more than 10% of the dataset represent dog breeds truly capture the human experience as a whole, or is it more reflective of dog owners' perspective? Similarly, is having a separate class for *"iPod"* — rather than a broader category like *"music player"* — an durable representation of the world?

### Problematic Groups

We categorize the problems with class selection into the following groups:

**Class Is a Subset/Special Case of Another Class**

- *"Indian elephant, Elephas maximus"* & *"African elephant, Loxodonta africana"* are also *"tusker"*
- *"bathtub, bathing tub, bath, tub"* is also a *"tub, vat"*

**Class Is a Part of Another Class Object**

- *"space bar"* is a part of *"computer keyboard, keypad"*
- *"car wheel"* wheel is a part of any vehicle class (*"racer, race car, racing car"*, *"sports car, sport car"*, *"minivan"*, etc.)

**Near Synonyms as Understood by Non-experts**

- *"laptop, laptop computer"* & *"notebook, notebook computer"*
- *"sunglasses, dark glasses, shades"* & *"sunglass"*

**Mostly Occur Together**

- *"sea anemone"* & *"anemone fish"*
- *"microphone, mike"* & *"stage"*

To identify such groups within the dataset, we conducted an analysis using the updated ImageNet training labels from the Re-labeling ImageNet paper [8:1], where an EfficientNet-L2 model was applied.

We performed hierarchical clustering on the classes with high EfficientNet-L2 error rates. Next, we manually reviewed the clusters, defined the mentioned categories, and organized the images accordingly. Each cluster consists of between 2 and 10 classes. Ultimately, this process led to the identification of 151 classes, which we organized into 48 groups. Each group contains a list of classes alongside their corresponding category. In some cases, multiple predefined relationships apply to the same classes, so a single group may span several categories.

The full list of problematic categories can be found here.

The OpenCLIP accuracies for both problematic and non-problematic groups of classes are given in **Table 3**.

| Dataset | Overall Accuracy | Problematic Classes Accuracy | Non-problematic Classes Accuracy |
| --- | --- | --- | --- |
| Validation | 84.61% | 73.47% | 86.59% |
| Training | 86.11% | 75.44% | 88.02% |

Table 3. OpenCLIP accuracy on ImageNet-1k.

The classes from the problematic groups significant lower OpenCLIP accuracy.

## Addressing Duplicates

Of all prior studies, **When Does Dough Become a Bagel** [9:2] examined the issue of duplicates most extensively. The paper identified 797 validation images that also appear in the training set, with some images occurring multiple times. They also highlighted the broader problem of near duplicates in ImageNet-1k (e.g. images from the same photoshoot). However, no statistics were provided since near duplicate are significanlty more difficult to detection than identical images.

The *"planetarium"* class mentioned earlier is a great example. It contains many near-duplicate images, as was noted in the [section focused on distribution shift](). Specifically, 68% of the validation images featured the same building in Buenos Aires. This observation naturally led us to investigate the issue of image duplicates more comprehensively.

Our analysis focuses on three types of duplicate sets:

1. **Cross-duplicates** between the validation and training sets (identified in earlier research).
2. **Duplicates within the validation set** (new findings).
3. **Duplicates within the training set** (new findings).

The search began with the duplicate candidate detection process (for more details, see [Appendix B]()). We then categorized duplicates in 2 groups: **exact duplicates** and **near duplicates**, and the results are surprising…

## Exact Duplicate Search: Pixel-Level Comparisons

Once duplicate candidates were identified, we conducted a pixel-wise comparison to classify exact duplicates. If two images had no pixel differences, they were marked as exact duplicates.

**Key Findings**

- In the **validation set**, 29 duplicate pairs were found. Each image in a pair belonged to a different ImageNet class.
- In the **training set**, 5,836 images were grouped into duplicates, with 2 to 4 images per group. Although most of these groups (5,724) contained images assigned to different classes, this highlights that class-based deduplication was not performed during the dataset's creation.
- For the **cross-validation-training search**, we discovered that 797 images in the validation set had duplicates in the training set. All these duplicate groups also consisted of images assigned to different ImageNet classes which is in agreement with previous studies [9:3].

**Bonus**

- In the **test set**, 89 duplicate pairs were found.

Since labels for the test set are not publicly available, we cannot determine whether the images in each pair have the same label or not. However, given that the test and validation sets were created simultaneously by splitting the collected evaluation data, we can infer that the situation is likely similar to the validation set. This suggests that each image in a pair belongs to a different class.

After finding exact duplicates, we removed them and recalculated accuracies of two models: OpenCLIP and an ImageNet-pretrained CNN EfficientNetV2. We conducted three experiments. First, we removed all duplicate pairs in the validation set. Next, we removed all duplicate images in the validation set that were also present in the training set (referred to as cross duplicates). Finally, we combined these two methods to remove all exact duplicates. In summary, our approach led to a 0.7% accuracy increase for the zero-shot model and a 1% accuracy increase for the pretrained CNN. We remind the reader that all exact duplicates have different labels and their erroneous classfication is very likely; the improvement is thus expected.

| Model | Overall | × Val | × Cross | × Val+Cross |
| --- | --- | --- | --- | --- |
| OpenCLIP | 84.61 | 84.67 | 85.27 | 85.32 |
| EfficientNetV2 | 85.56 | 85.62 | 86.51 | 86.57 |

Table 4. OpenCLIP and EfficientNet accuracies on the whole ImageNet-1k (overall) and without different kinds of *exact duplicates*.

## Near Duplicate Detection Method

The initial automatic search for duplicates was followed by a careful manual review of duplicate candidates images. After the review, each image was classified into one of the following near-duplicate groups.

**Image Augmentations**: images that result from various transformations applied to an original image, such as cropping, resizing, blurring, adding text, rotating, mirroring, or changing colors. An example is shown below.

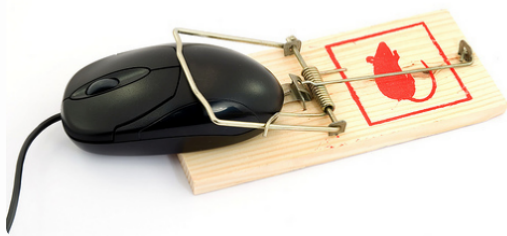
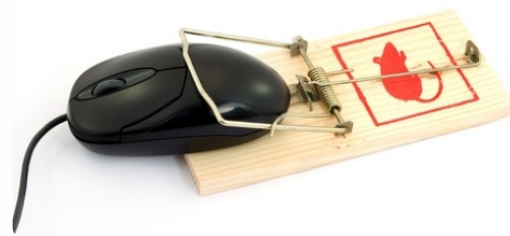

(a) *"computer mouse"* from the validation set

(b) *"mousetrap"* from the training set

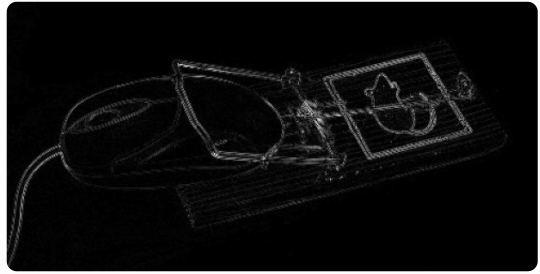
(c) the difference of (a) and (b)

Figure 9. Near duplicates - Image Augmentation.

**Similar View**: images of the same object taken from slightly different angles at different times. An example is depicted below.

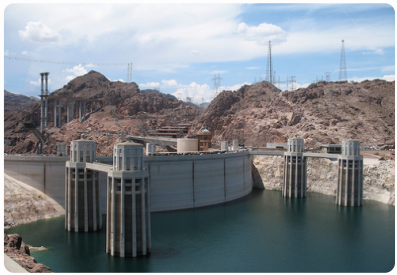
*"dam, dike, dyke"* from the validation set

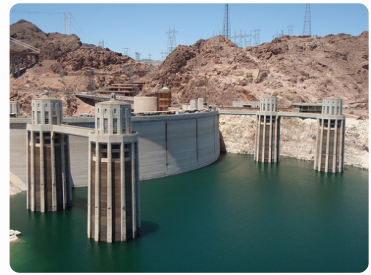
*"dam, dike, dyke"* from the training set

Figure 10. Near duplicates - Simalar View.

**Key Findings**

- In the **validation set**, 26 near-duplicate groups were found, involving 69 images in total. All duplicates in a groups had consistent labels, which helps maintain label reliability for model evaluation.
- For the **cross-validation-training search**, we discovered that 269 images from the validation set matched 400 training images.

We continued evaluating models with near duplicates removed. First, we removed all near duplicate groups in the validation set. Next, we removed validation images that appeared in the training set (referred to as near cross duplicates), then we removed both. Lastly, we removed all exact duplicates and near duplicates from the validation set. As shown in **Table 5**, removing near duplicates had minimal impact on accuracy, as these images were mostly consistently assigned the same label within each duplicate group.

| Model | Overall | × Val | × Cross | × Val+Cross | × All |
|---|---|---|---|---|---|
| OpenCLIP | 84.61 | 84.60 | 84.63 | 84.62 | 85.32 |
| EfficientNetV2 | 85.56 | 85.54 | 85.59 | 85.59 | 86.59 |

Table 5. OpenCLIP and EfficientNet accuracies on the whole ImageNet-1k (overall) and without different kinds of duplicates.

## Prompting Vision-Language Models

Issues with dataset construction, such as overlapping or imprecise WordNet synsets that may evolve over time, raise questions about their impact on evaluation of vision-language models like CLIP [12:1].

CLIP zero-shot classification is based on the distance of the image embeddings to the text embeddings representing each class. A natural approach is to create class embeddings based on the WordNet synset names. However, there are issues.

As mentioned in the section about dataset construction, WordNet synset names typically consist of multiple terms. For example, the term maillot appears in both *"maillot"* and *"maillot, tank suit"*. The first synset definition is "tights for dancers or gymnasts", while the second one is "a woman's one-piece bathing suit". This can create significant difficulties for any VLM.

An investigation of the CLIP codebase reveals the authors, created a customized OpenAI version of the ImageNet-1k class names.
Despite not being explicitly stated in the original work, it seems the authors were aware of many of the issues in the original class names.
In the notebook, the authors suggest further work with class names is necessary, a sentiment we agree with.

**Class Text Prompt Modifications: An Empirical Study**

To illustrate the impact of class names in zero-shot recognition, we devloped a new set of "modified" class names, building on OpenAI's version. In the experiments, we decided to use OpenCLIP, an open-source implementation that outperforms the original CLIP model.

**Table 6** shows recognition accuracy for the five classes with the most significant gain when using OpenAI class names vs. the original ImageNet names. The changes of the text whose embedding is used primarily address CLIP's need for broader context. For instance, in ImageNet, *"sorrel"* refers to a horse coloring, while in common usage, we're used to hearing it refer to a plant. This can be a problem for VLMs due to the lack of context, which in turn the new class name *"common sorrel horse"* provides.

| ImageNet Class Name (WordNet) | | | | OpenAI Class Name |
|---|---|---|---|---|
| *"sorrel"* | | 0% | 98% | *"common sorrel horse"* |
| *"bluetick"* | | 0% | 78% | *"Bluetick Coonhound"* |

| | | | |
|---|---|---|---|
| "redbone" | 0% | 78% | "Redbone Coonhound" |
| "rock beauty, Holocanthus tricolor" | 22% | 96% | "rock beauty fish" |
| "notebook, notebook computer" | 16% | 66% | "notebook computer" |

Table 6. OpenCLIP zero-shot recognition accuracy with ImageNet (left) and OpenAI text prompts (right).

**Table 7** demonstrates the improvement of our modifications w.r.t. OpenAI's class names. Notably, renaming *"coffee maker"* to *"coffeepot"* not only increased accuracy within this class but also positively impacted the class *"espresso machine"*, where no changes were made.

| OpenAI Class Name | | | "Modified" Class Name |
|---|---|---|---|
| "canoe" | 48% | 100% | "kayak" |
| "vespa" | 42% | 82% | "motor scooter" |
| "coffeemaker" | 48% | 84% | "coffeepot" |
| "sailboat" | 76% | 100% | "yawl (boat)" |
| "espresso machine" | 50% | 72% | "espresso machine" |

Table 7. OpenCLIP zero-shot recognition accuracy with OpenAI (left) and text prompts "modified" by us (right). The "canoe" ImageNet-1k class achieves 100% accuracy if prompted by 'kayak'. This is not surprising, given that *all images in the "canoe" vladiation set depict kayaks*. There is no 'kayak' class in ImageNet-1k.

Our modifications were found by trial and error, which suggests that there is a large space for possible improvement in VLM text prompting.

## Fixing ImageNet Labels: A Case Study

Do you know the precise difference between a weasel, mink, polecat, black-footed ferret, domestic ferret, otter, badger, tayra, and marten? Most likely not. We use these animal species to illustrate the complexity of image labeling in ImageNet-1k. We enlisted an expert to help.

We consider images from the following classes:

- *"weasel"*
- *"mink"*
- *"polecat, fitch, foulmart, foumart, Mustela putorius"*
- *"black-footed ferret, ferret, Mustela nigripes"*

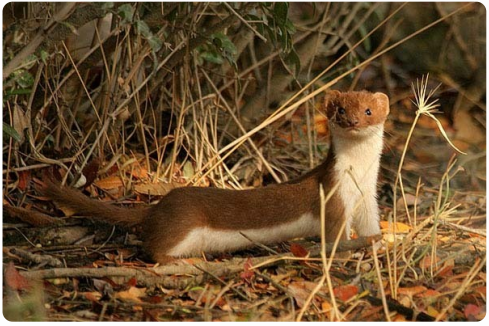
*"weasel"*

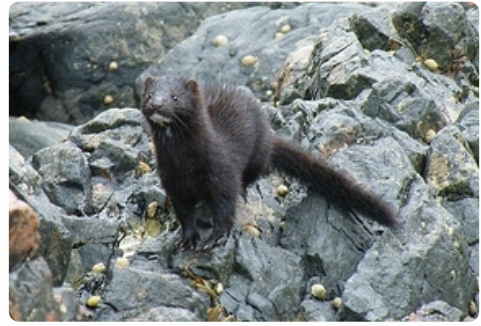
*"mink"*

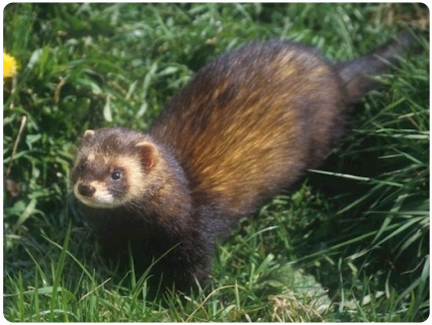
*"polecat, fitch, foulmart, foumart, Mustela putorius"*

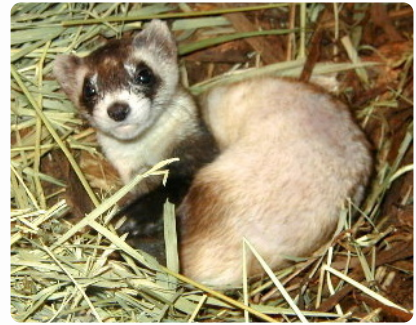
*"black-footed ferret, ferret, Mustela nigripes"*

Figure 11. Correctly labeled ImageNet-1k images from the *"weasel"*, *"mink"*, *"polecat"*, and *"black-footed ferret"* classes. The ground truth labels for these categories are often wrong, mainly confusing these species.

These classes have a high percentage of incorrect ground truth labels, both in the training and validation sets. Most of the errors are caused by confusion between the four classes but the sets also contain images depicting animals from other ImageNet-1k classes, such as otter or badger, as well as images from classes not in ImageNet-1k, e.g. vole or tayra. But that is not the sole issue.

## The Weasel Problem

Let us look at *"weasel"* class definitions:

- **WordNet**: 'small carnivorous mammal with short legs and elongated body and neck'.
- **Wikipedia**: 'The English word weasel was originally applied to one species of the genus, the European form of the least weasel (*Mustela nivalis*). This usage is retained in British English, where the name is also extended to cover several other small species of the genus. However, in technical discourse and in American usage, the term weasel can refer to any member of the genus, the genus as a whole, and even to members of the related genus *Neogale*''.
- **Webster** (broader): 'any of various small slender active carnivorous mammals (genus *Mustela* of the family *Mustelidae*, the weasel family) that are able to prey on animals

(such as rabbits) larger than themselves, are mostly brown with white or yellowish underparts, and in northern forms turn white in winter'.

The definition of the *"weasel"* synset in WordNet is too broad - it potentially encompasses all the other mentioned classes. Moreover, the interpretation of the term weasel varies, between UK and US English, further complicating its consistent application. In US English, the term weasel often refers to the whole *Mustelidae*, also called 'the weasel family'. All of the following - weasel, mink, European polecat, and black-footed ferret - belong to the weasel family, as understood by US English.

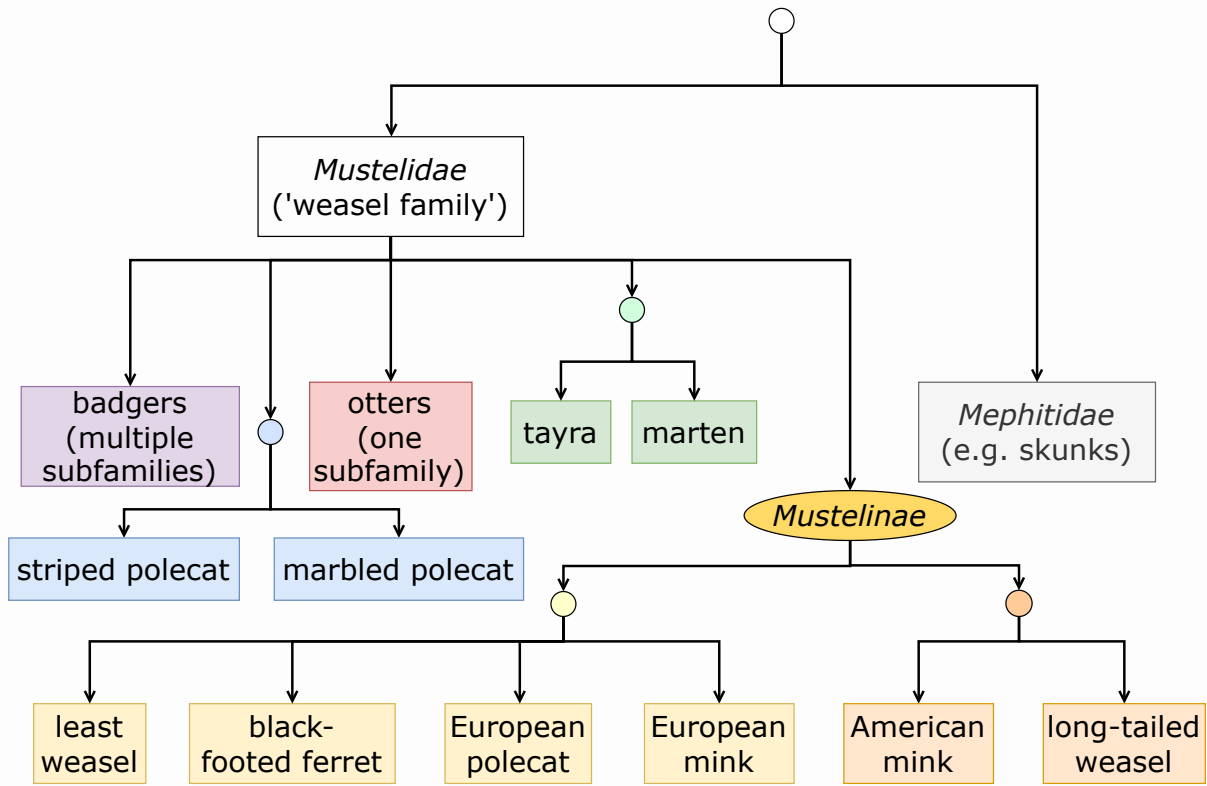

Figure 12. A simplified branching diagram showing the *Mustelidae* family, also referred to as 'the weasel family', or simply 'weasels' in US English. Some of ImageNet classes are: *"weasel"*, *"polecat"*, *"black-footed ferret"* and *"mink"*. All of these belong to the weasel family, pointing at yet another issue with class names. Images of these classes also depict species outside of ImageNet, such as tayras or voles, or even different ImageNet classes, such as otter or badger. The diagram shows evolutionary relationships between selected species from the ImageNet dataset; the circles mark higher taxonomic units.

### Solution

One possible solution is to define the *"weasel"* class more precisely as the subgenus *Mustela*, which contains the 'least weasel' and other very similar species, which would lead only to removal of a few images.

## The Ferret Problem

Another complication arises with the *"black-footed ferret"* and *"polecat"* classes:

- **WordNet synset name:** *"black-footed ferret, ferret, Mustela nigripes"*
- **ferret, Webster:** 'a domesticated usually albino, brownish, or silver-gray animal (*Mustela furo* synonym *Mustela putorius furo*) that is descended from the European polecat'.
- **black-footed ferret, Wikipedia:** 'the black-footed ferret is roughly the size of a mink and is similar in appearance to the European polecat and the Asian steppe polecat'.

The synset *"black-footed ferret, ferret, Mustela nigripes"* includes both the term 'black-footed ferret' and 'ferret'. The latter refers to a domesticated variety of the European polecat.

Consequently, the term 'ferret' is ambiguous; it may be understood both as a synonym for the black-footed ferret or as the domesticated polecat. Additionally, the domestic ferret and European polecat are nearly indistinguishable to non-specialists; even experts may face difficulties because these species can interbreed.

There is also a potential for contextual bias in labeling, as ferrets are commonly found in domestic environments or in the presence of humans.

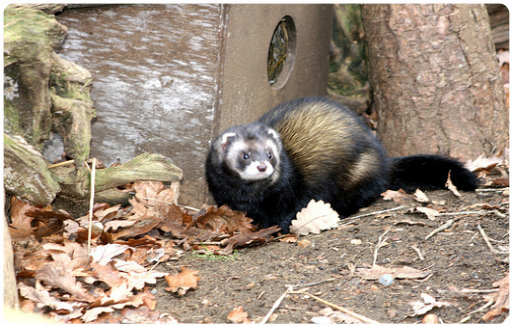
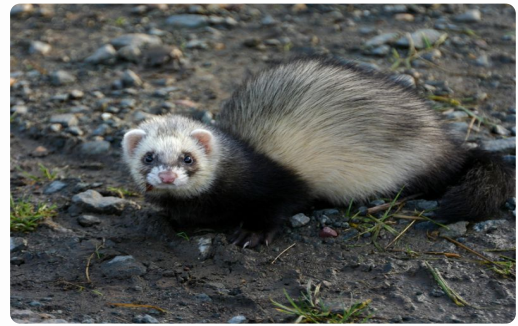

European polecat	domestic ferret

Figure 13. Training set images from the ImageNet-1k *"polecat"* class. Distinguishing between European polecat and domestic ferret is challenging due to the similarity in appearence. European polecats tend to be more muscular than domestic ferrets, have overall darker fur and well-defined white face mask.

To make matters worse, in the **validation set for the class *"black-footed ferret"*, only one image depicts this species!** A solution to this problem thus requires not only removal, or transfer to the correct class, of the incorrectly labeled images, but also collection of new data.

The term polecat presents a similar ambiguity w.r.t. the term ferret, as it is currently included in two synsets. One synset refers to skunk (family *Mephitidae*), while the other to *Mustela putorius*, the European polecat. These are in line with the definitions of the word polecat here.

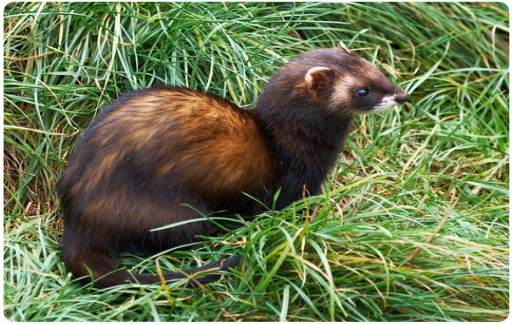
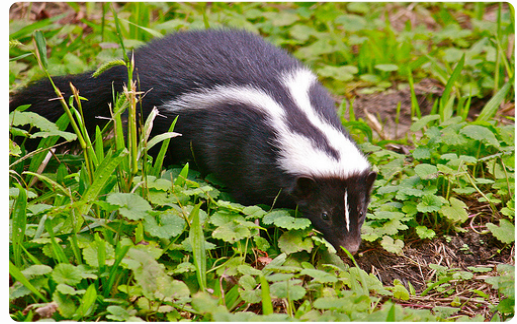

European polecat           skunk

Figure 14. European polecat and skunk. These species share the term 'polecat' in their synsets but belong to different families (see Figure 12).

**Solution**

To solve the 'ferret' issues, redefinition of classes might be needed, e.g.:

1. Introduction of a distinct class for ferret, specifically denoting the domesticated form of the European polecat.
2. Reclassification of the term polecat so that it no longer appears in the synset for skunk; instead, this term should be used to represent a broader category encompassing both the European polecat and American polecat (also referred to as the back-footed ferret), as well as other species, such as the marbled, steppe, and striped polecats.
3. Create a class that encompasses both polecats and ferrets.

## Results of Relabeling

After relabeling the weasel family classes, we find that only the *"mink"* class had more than 50% of labels correct.
The percentage of the **correctly** labeled images in ImageNet-1k was:

| Class Name | Percentage Correctly Labeled |
| --- | --- |
| Weasel | 44% |
| Mink | 68% |
| Polecat | 32% |
| Black-footed ferret | 2% |

The misclassified images either show an animal from the aforementioned classes, or from a different ImageNet class (such as otter or badger). There are also images of animals outside of ImageNet-1k classes, while some images are ambiguous, see Figure 15.

Images of animals that do not belong to any ImageNet class are assigned to the 'non-ImageNet' label in the graph shown in Figure 16. This category includes animals such as vole, tayra, marten, and Chinese ferret-badger. Although 'domestic ferret' is also a non-ImageNet label, it is shown separately because of its large representation in the sets.

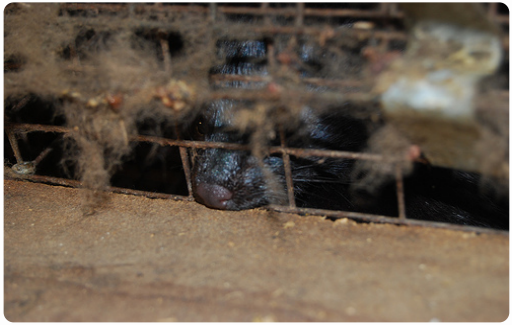
characteristic features obscured

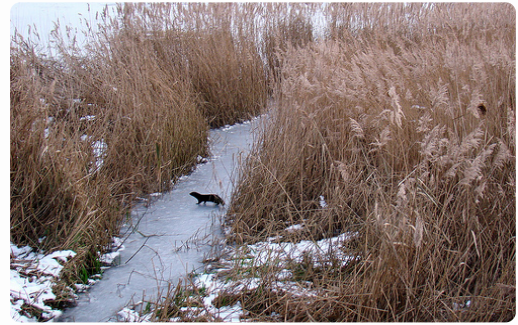
too great a distance for identification

Figure 15. Images from the validation set *"mink"* class, which our expert labeled as ambiguous. Left: the characteristic features are obscured; most likely a mink, but possibly a species from genus *'Martes'*. Right: an animal from genus *'Mustela'*; possibly a mink or another dark-furred species.

The 'ambiguous' label is used for images that are blurry, have the characteristic features of the species obscured, show the species from too great a distance, or have other flaws that prevent unequivocal identification of the species.

Let us take a closer look at the four examined classes.

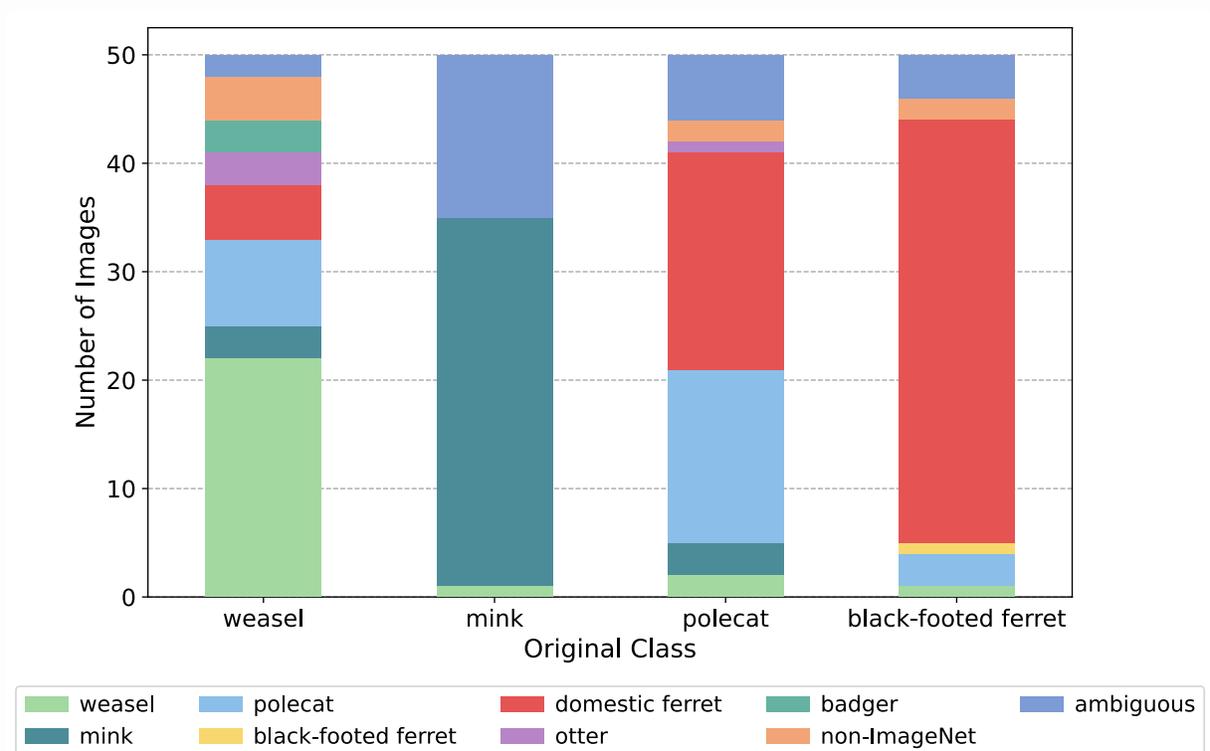

Figure 16. Validation set relabeling results for the *"weasel"*, *"mink"*, *"polecat, fitch, foulmart, foumart, Mustela putorius"*, and *"black-footed ferret, ferret, Mustela nigripes"* classes.

The weasel class contains a wide variety of misclassified images. This includes minks (6%), polecats (16%), domestic ferrets (10%), otters (6%), and badgers (6%). The high rate of misclassification may be due to the unclear definition of this class, as all of these species belong to the weasel family discussed earlier.

The *"mink"* class is predominantly correctly labeled but a substantial portion (30%) of images is ambiguous; meaning they are low quality or the subject is obscured. These images should preferably be removed or assign multiple possible labels (single object but ambigous).

The *"polecat"* class has a significant percentage (40%) of images depicting domestic ferrets. That is not surprising as distinguishing between polecats and domestic ferrets is particularly challenging.

Finally, the *"black-footed ferret"* class contains only one image of this species, while the majority (80%) of the images depict domestic ferrets.

Luccioni and Rolnick (2022) [13] analyzed the classes representing wildlife species and the misrepresentation of biodiversity within the ImageNet-1k dataset. Their findings reveal a substantial proportion of incorrect labels across these classes. Notably, they examined the class *"black-footed ferret"* and reported results consistent with those observed in our relabeling process.

## Conclusion

We presented a number of problems, some known and some new, of the ImageNet-1k dataset. The blog mainly focuses on a precise description of the issues and their "size", i.e., what fraction of the image-label pairs it affects. In some cases, we discuss solutions. We hope that one of the outcomes of publishing this analysis is that it will open a broader exchange of ideas on what is and what is not worth fixing. Every solution will involve trade-offs, at least between correctness and backward compatability. The wide use of the dataset makes it difficult to assess the impact of any change; possibly we are heading towards multiple evaluation protocals and ground-truth versions.

The activity opened many questions. First and foremost: "Is this worth it?" and "Will the community benefit from (much) more accurate ImageNet-1k (and other commonly used sets) re-labeling and class definitions?". For the validation set, which is used for performance evaluation of a wide range of models, the answer seems a clear "yes". For the training set, we see benefits too. For instance, it seems that learning a fine-grained model from very noisy data is very challenging.

The answers to the questions above depend on the effort needed to re-label the images and to clean the class definitions. Our experience is that current tools, both state-of-the-art classifiers and zero-shot VLM model reduce the need for manual effort significantly.

The effort to find precise, unambigous definitions of the ImageNet-1k classes lead us to the use of VLMs and LLMs. The LLM responses were accurate and informative, if prompted properly, even warning about common causes of confusion. It seems that LLMs are very suitable for annotator training.

In fact, VLMs might not only be a useful tool in this context, but their performance might improve if a large accurately labeled dataset is available. A joint development of ImageNet and WordNet is desirable, as the problems with class definitions attest.

We hired an expert annotator in order to obtain precise annotation of the weasel-like animal classes analyzed in the case study.
Expert annotators help identifying subtle nuances and complexities within the dataset that might be easily overlooked by non-specialists. On the other hand, their understanding of certain terms might not coincide with common usage. We might need parameterizable VLM models, e.g., for professional and technical use as well as for the vernacular.
In prior work [7:1], MTurkers have been used to find incorrect labels. However, we found that they missed many problems. These errors are correlated, and attempts to removed them, by e.g. majority voting, cannot detect them. When it comes to highly accurate labels, an expert is worth not a thousand, but any number of MTurkers.

Some of the issues, like the presence of image duplicates and near duplicates, may create an opportunity for performing meta-experiments. For instance, what if two methods with identical overall performance on ImageNet-1k differ significantly in accuracy on duplicates? What is the interpretation of a situation where a method performs well on classes with many incorrectly labeled images?

In many areas of computer vision, models reached accuracy comparable to the so-called ground truth, losing the compass pointing to better performance. As we have seen, improving ground truth quality is not a simple task of checking and re-checking, but touches some core issues of both vision and language modeling. This blog is a small step towards resetting the compass for ImageNet-1k.

# Appendix A

## A.1. Previous Work in Detail

**Imagenet Multilabel** [4:1] The authors reannotated a class-balanced subset of images, covering 40% of the validation set. The main focus was on assigning multiple labels to capture the presence of several objects from different ImageNet classes. The work highlighted the extent of the multilabel problem.

**Contextualizing Progress on Benchmarks** [6:1] The authors reannotated 20% of validation images from each class with predictions from an ensemble of deep models, which were verified by non-expert annotators.
Their analysis revealed over 21% images were multilabel. The work also lists groups of most commonly co-occurring classes, like "race car" and "car wheel" (see Figure 8).

**Imagenet Real** [5:1] aimed at correcting all labels in the validation set. The original label was deemed correct if it agreed with predictions of all 6 deep models selected by the authors. This was the case for approximately 50% of images. For the remaining images, labels were manually reviewed by five human annotators.
About 15% of the images are reported multilabel and about 6% as ambiguous (which were left unlabeled).
The training set was checked using BiT-L [11:1] model predictions, without human involvement. In 10-fold cross-validation, approximately 90% of the training images had labels consistent with model predictions.

**Label Errors** [7:2] used Confident Larning framework [14] and five Amazon Mechanical Turk (MTurk) workers for re-annotation. Images lacking workers' consensus — about 11% — were marked as "unclear".
The label error on the validation set is estimated to be about 20%.

**Re-labelling ImageNet** [8:2] authors used the EfficientNet-L2 [15] model trained on the JFT-300M dataset to reannotate the training set and to convert ImageNet's single-label annotations into multilabel annotations with localized information.

**When Does Dough Become a Bagel?** [9:4] The study identifies duplicate images in the validation and training sets. Interestingly, each image in a group of duplicates is found to have a different label. This indicates duplicates were removed only within the same class, incorrectly assuming that duplicate images cannot have different label. The work emphasizes the importance of addressing not only duplicates but also near-duplicate images, e.g., similar photos from the same photoshoot.

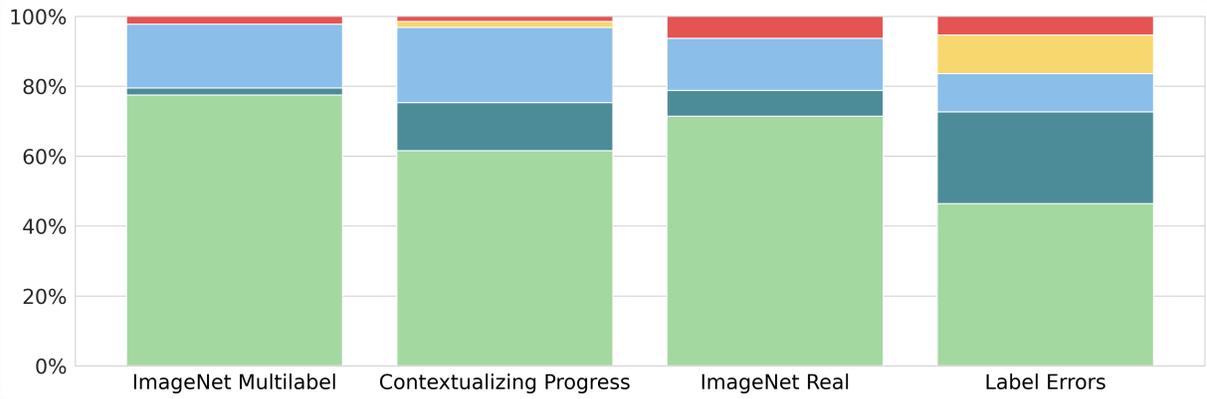

Figure 17. Error analyses for the ImageNet validation set, presented separately for each study discussed in this section. ■ Single label image, original label is correct. ■ Single label image, original label is incorrect, full agreement on correction. ■ Multilabel images. ■ Single label image, inconsistent label corrections. ■ Ambiguous, no agreement on the label.

### A.2. Results Evaluation Note

It should be noted that for the **Label Errors** [7:3] paper (see Previous Work in Detail), the .json file containing Mturk decisions was used and evaluated with a modification from the original methodology. Instead of using the majority consensus (3+ out of 5 workers), only decisions unanimously agreed upon by all 5 workers were considered.

# Appendix B

**Initial Approach: DINOv2 Over CLIP**

We computed image embeddings using the DINOv2 [16] model. We initially considered using CLIP for this, but its results were not satisfactory. Once the embeddings were generated, we applied the *K-Nearest Neighbors (K-NN) algorithm* to detect possible duplicates based on their similarity in the embedding space.

**Duplicate Candidate Detection Method**

The algorithm checks how close the embeddings of two images are. If the distance between them is less than a certain threshold (confidence level), we marked them as possible duplicates.

Let us break it down:

- Each image $I_i$ has an embedding, $\mathbf{e}(I_i)$, in the feature space.
- The *K-NN* algorithm finds the 5 closest neighbors for each image.
- $d(\mathbf{e}(I_i), \mathbf{e}(I_j))$ represents the cosine distance between the embeddings of two images.
- $\tau$ is a predefined confidence threshold chosen high enough to ensure that no true positives are lost.

Mathematically, the condition for possible duplicates is:

$$d(\mathbf{e}(I_i), \mathbf{e}(I_j)) \leq \tau$$